\documentclass{article}
\usepackage{kr}

\usepackage{times}
\usepackage{soul}
\usepackage{url}
\usepackage[utf8]{inputenc}
\usepackage[small]{caption}
\usepackage{graphicx}
\usepackage{amsmath}
\usepackage{amsthm}
\usepackage{booktabs}
\usepackage{algorithm}
\usepackage[letterpaper,top=2cm,bottom=2cm,left=3cm,right=3cm,marginparwidth=1.75cm]{geometry}
\usepackage{amsmath,amsthm}
\usepackage{physics}
\usepackage{graphicx}
\usepackage{float}
\usepackage[colorlinks=true, allcolors=blue]{hyperref}

\usepackage[english]{babel}
\usepackage{setspace}
\usepackage{tikz}
\usetikzlibrary{automata,topaths}
\usepackage{xcolor}
\usepackage{bbm}
\usepackage{amsthm,amsmath,amsfonts}
\usepackage{algorithm}
\usepackage{algpseudocode}

\newtheorem{theorem}{Theorem}

\newtheorem{remark}[theorem]{Remark}

\newtheorem{proposition}[theorem]{Proposition}

\def\R{\mathbb{R}}

\def\Ebb{\mathbb{E}}

\def\Lcal{\mathcal{L}}

\def\Xcal{\mathcal{X}}
\def\Ycal{\mathcal{Y}}
\def\Zcal{\mathcal{Z}}

\def\ybf{\mathbf{y}}

\def\zbf{\mathbf{z}}

\title{Deep Conditional Measure Quantization}
\author{Gabriel Turinici 
	\affiliations
	CEREMADE, Universit\'e Paris Dauphine - PSL, Paris, France
    \emails
Gabriel.Turinici@dauphine.fr    
\\ \ \\ 
\today
}

\date{\today}

\begin{document}
	\maketitle
	
\begin{abstract}
Quantization of a probability measure means representing it with a finite set of Dirac masses that approximates the input distribution well enough (in some metric space of probability measures). 
Various methods exists to do so, but the situation of quantizing a conditional law has been less explored. We propose a method, called DCMQ, involving a Huber-energy kernel-based approach coupled with a deep neural network architecture. The method is tested on several examples and obtains promising results.
\end{abstract}
	
\section{Introduction}
	
\subsection{Conditional measure quantization : motivation}

In general terms, quantization is the process of replacing a set of values (possibly an infinity of them) with a finite number chosen to be the most representative according to some metric. 
This is related to vector quantization \cite{book_quantization_measures,measure_quantization}
that operate on objects in a high dimensional space. 
Applications range from signal processing~\cite{constantinides1984quantization,crochiere1989multirate,vaidyanathan1993multirate}
to finance~\cite{lyons1995cubature,pages_optimal_2018,pages2015cubature}, 
statistics~\cite{sculley2010web,optimal_quantization_wasserstein11}, which makes it an important area of research in statistics, knowledge representation and machine learning. We will be concerned with a particular instance of this question, namely the quantization of probability measures\footnote{We only consider here probability measures, but the extension to signed measures can be done directly following the prescriptions in \cite{huber_energy_quantization_turinici22}.}; in this case we want to {represent the knowledge encoded into} a probability measure $\mu$ with support in $\Ycal$ by a sum of $Q$ Dirac masses 
	$\delta^\beta_\ybf= \sum_{q=1}^Q \beta_q \delta_{y_q}$, where $y_q \in \Ycal$ are chosen such that
	the distance between $\mu$ and $\delta^\beta_\ybf$ is as small as possible (we will come back later to the definition of the distance); here $\beta$ are some weight parameters (see section ~\ref{sec:theory}).
	
	But, there are times when $\mu$ is depending itself on another parameter $\mu=\mu_x$ or $\mu$ can be a conditional law. The main question treated in this paper is how to compute efficiently the quantization $\mu_x$ of the law $\mu$ conditional to $x$. Our proposal is to involve a deep neural network that minimizes  the Huber-energy statistical distance (see~\cite{huber_energy_quantization_turinici22} for a definition) and outputs the quantized version of the conditional law.
	
	The outline of the paper is the following: in the rest of this section we recall some  related works from the literature; we present some theoretical information in section~\ref{sec:theory}; the practical implementation of the method is described in sections~\ref{sec:algo} and \ref{sec:numerics_mnist_reconstruction} together with numerical results. Concluding remarks are the object of section~\ref{sec:concluding}.

	\subsection{Brief literature review on measure quantization, conditional sampling and conditional quantization}

In the general area of (non-conditional) measure quantization, a related proposal~\cite{chatalic_nystrom_2022} investigates the 
Nystrom mean embeddings, that constructs the quantization based on the exploitation of a small random subset of the dataset.

	The literature on general conditional quantization is very scarce, but some works have been done in connection to the  so called 'conditional $L^1$ median~\footnote{The $L^1$ median of a measure $\xi$ (with finite first order moment) is the point that minimizes, with respect to $y$, the first order moment of $\xi$ centered at $y$ (the median does so in one dimension and can be extended by this definition to several dimensions.}. 
	Several works 
	such as~\cite{berlinet_estimation_2001,berlinet_conditional_2001}
	ask the question of the conditional $L^1$ median; our approach is similar to that one, with the difference that we look for a general quantization, not only a quantization with a single ($Q=1$) point and we are not attached to the first order moment. See also 
	\cite{white_nonparametric_1992}
	for a proposal involving neural network computations of conditional quantiles.

In a related work, Zhou et al. \cite{zhou_deep_2021}
propose a generative approach to sample from a conditional distribution by learning a conditional generator. They exploit a Kullbak-Liebler (KL) divergence formulation 
(in \cite{cond_energy_distrib} a 'energy' kernel is used instead) 
but the generator itself is not quantized.

On the contrary, Vuong et al. \cite{vuong_vector_2023} learn a deep discrete representation using the Wasserstein distance but their approach is not targeted towards conditional distribution representation.

In signal processing, conditional quantization goes often by the name of 
conditional vector quantization and has been used e.g. 
for speech encoding in~\cite{agiomyrgiannakis_conditional_2007}, see also
\cite{parks1985chebyshev,proakis1995digital,graham1972efficient} for other references in the signal processing area. Our approach differs by choosing to treat the question in a general, not application dependent way which materializes into the choice of the Huber-energy statistical distance and the use of deep neural networks (hereafter called 'DNN') for interpolation.

Natural language processing is also an application domain; for instance, even if neither 
GPT-3 \cite{brown2020gpt3,gpt_ref_arxiv}~\footnote{short for "Generative Pre-training Transformer 3", the large language model developed by OpenAI.} nor its 
follow-up ChatGPT \cite{noauthor_chatgpt_2022} are designed explicitly as quantizers, in practice ChatGPT will answer based on a user question or previous conversation. Put it otherwise, it selects a small set of possible answers from the conditional probability of any text given previous contents. This is indeed a conditional quantization.

\subsubsection{Interpolation by deep neural networks}
	
The conditional quantization, or quantization depending on some parameter, can also be viewed as some kind of interpolation. Given a set of parameters $x_1,...,x_L$ we can pre-compute the  
quantized conditional distribution for these parameters and then, for any new choice of the parameter, interpolate using the precomputed data.
	
	Several works explored the use of DNNs for interpolation, for instance in~\cite{Wang_2019_CVPR} a DNN is trained to learn a mapping from input data points to output data points; then, at the prediction time the DNN can generate an interpolated value for any intermediate input value; the applications range from image processing~\cite{Wang_2019_CVPR} to audio interpolation~\cite{suefusa_anomalous_2020}; in \cite{zhou2020comparison}
	data interpolation was used in scientific simulations such as weather forecasting and fluid dynamics simulations. 
	
	Other approaches to using DNNs for interpolation involve using the DNN to learn a probabilistic model of the data~\cite{astonpr373}, and generate the interpolated values using the learned data distribution (with applications to natural language processing and time series analysis).  In NLP the possibility 
	of language models to learn to infill (missing parts of) text~\cite{bavarian_efficient_2022} can also be considered close to a extrapolation method.
	
	\section{The deep neural network conditional quantization method}
	\label{sec:theory}
	
	\subsection{Setting and notations}
	
	We follow the usual notation with lowercase letter for values, upper case for random variables and bold face for vectors. Let $\Xcal = \R^{n_x}$, $\Ycal = \R^{n_y}$ (with $n_x$, $n_y$ non-null integers)~\footnote{What is said here can be extended to the situation when 
		$\Xcal$ or $\Ycal$ are only open subsets of $\R^{n_x}$ or $\R^{n_y}$.}
	and $\mu$ a joint law with support in $\Xcal \times \Ycal$. Denote 
	$\mu^X$ and $\mu^Y$ the marginals of the law $\mu$; for instance, 
	if $X$ and $Y$ are two random variable with support in $\Xcal$ and $\Ycal$ respectively, and if $(X,Y)$ follows the law $\mu$ then $\mu^X$ is the law of $X$ and $\mu^Y$ is the law of $Y$.
	We look for a method to quantize the distribution 
	\begin{equation}
		\mu_x = \mu(dy | X=x),
	\end{equation}
	of $Y$ conditional to $X=x$.
	Fixing an integer $Q > 0$, and some weights $\beta \in \R^Q$ (that sum up to one) we look for $\ybf(x) \in \Ycal^Q$ such that 
	$\delta_{\beta,\ybf(x)}= \sum_{q=1}^Q \beta_q \delta_{\ybf_q(x)}$ 
	is as close as possible to $\mu_x$.
	
	To describe what 'close' means, we need to use a distance $d$ defined on the set of probability measures; the distance we use will be the Huber-energy distance that we define below~\cite{szekely2005hierarchical,szekely_energy_2013,huber_energy_quantization_turinici22}; given $a\ge0$ the Huber-energy (negative definite) kernel is defined by $h_{a,r}(z,\tilde z)=
	(a^2+ |z- \tilde z|^2)^{r/2} - a^r$
	( $a\ge0$, $r\in ]0,2[$)~; 
	it is known (see \cite{sriperumbudur2010hilbert,huber_energy_quantization_turinici22} and related works) that $h$ induces a distance $d$~: 
	for any two probability laws $\eta$, $\tilde{\eta}$ on some space 
	$\Zcal$~\footnote{Here $\Zcal$ will be either $\Xcal \times \Ycal$ or $\Ycal$.} with $\int_\Zcal |z|^r\eta(dz) < \infty$, 
	$\int_\Zcal |z|^r\tilde\eta(dz) < \infty$, 
	we can write~:
	\begin{equation}
		d(\eta,\tilde\eta)^2 = -\frac{1}{2}\int_\Zcal \int_\Zcal  h_{a,r}(z,\tilde{z}) (\eta-\tilde\eta)(dz) (\eta-\tilde\eta)(d\tilde{z}).
		\label{eq:defdistance}
	\end{equation}
	Note that in particular $h_{a,r}(z,\tilde{z})= d(\delta_{z},\delta_{\tilde z})^2$ and moreover
	we have for $\zbf\in \Zcal^L, \beta \in \R^L$, 
	$\tilde \zbf \in \Zcal^M, \tilde \beta \in \R^M$,
	\begin{eqnarray} & \ & \!\!\!\!\!\!\!\!\!\!\!\!\!\!\!\!\!\!\!\!\!
		d(\delta^\beta_\zbf,\delta^{\tilde\beta}_{\tilde \zbf})^2 = 
		\sum_{i=1}^L \sum_{j=1}^M \beta_i \tilde\beta_j h_{a,r}(z_i,\tilde z_j)
\nonumber \\ & \ & \!\!\!\!\!\!\!\!\!\!\!\!\!\!\!\!\!\!\!\!\!\!\!\!\!\!\!\!\!\!
		-\frac{1}{2} \sum_{i,j=1}^L \beta_i \beta_j h_{a,r}(z_i,z_j)
		-\frac{1}{2} \sum_{i,j=1}^M \tilde\beta_i \tilde\beta_j h_{a,r}(\tilde z_i,\tilde z_j).
		\label{eq:defdistance_dirac}
	\end{eqnarray}
	In this paper we will only be concerned with uniform weights (but what is said here can be extended to arbitrary, but fixed, weights, see \cite{huber_energy_quantization_turinici22} for related considerations);
	in this case we write simply~:
	\begin{equation}
		\delta_{\zbf}= \frac{1}{L} \sum_{\ell=1}^L  \delta_{z_\ell}.
	\end{equation}
	
	The conditional quantization of the law $\mu$ is defined as follows: for any $x\in \Xcal$ we look for the minimizer
	of the distance to $\mu_x$
 i.e., any 	 $\ybf^{opt}(x) \in \Ycal^Q$ such that~: 
	\begin{equation}
		d(\delta_{\ybf^{opt}(x)},\mu_x)^2 \le 
		d(\delta_{\ybf},\mu_x)^2,~~\forall \ybf \in \Ycal^Q.
		\label{eq:defintion_quantizer_by_ineq}
	\end{equation}
	Note that in general the minimum is not unique so $\ybf^{opt}(x)$ is a set-valued function. Accordingly a first theoretical important question is whether one can find a selection~\footnote{A selection of a set-valued function $g:A \to 2^B \setminus \emptyset$ is a function $\tilde g: A \to B$ such that 
$\forall a \in A : \tilde g (a) \in g(a)$. Recall that $2^B$ is the set of all subsets of $B$.} of $\ybf^{opt}(x)$ with good properties such as measurability, continuity etc. These questions are answered in the rest of this section. Then a practical question is how to find convenient conditional quantizations; this is described in sections~\ref{sec:algo} and \ref{sec:numerics_mnist_reconstruction}.
	
	\subsection{Existence of a measurable conditional quantization}
	We answer here the question of whether there exists a proper measurable function (i.e. not a set valued function) that represents the conditional quantization.  The answer is given in the following~:
	\begin{proposition} 
		Suppose that the bi-variate distribution $\mu$ is such that for any $x\in \Xcal$ the distribution $\mu_x$ has finite $r$-th order moment. Then there exists a measurable function $\ybf^{opt} : \Xcal \to \Ycal^Q$ such that
		$\ybf^{opt}(x)$ 
		satisfies  equation
		\eqref{eq:defintion_quantizer_by_ineq} for any $x\in\Xcal$.
		\label{prop:measurable} 
	\end{proposition}
	\begin{remark}
		The existence of the $r$-th order moment condition is only required because of our choice of kernel and in general can be weakened.
	\end{remark}
	\begin{proof}
		A general proof
		can be obtained using the 
		Kuratowski and Ryll-Nardzewski measurable selection theorem \cite{kuratowski_general_1965} (see also \cite{cascales_measurability_2010})
		but we will follow the faster route that employs the Corollary~1 in \cite[page 904]{brown_measurable_1973}. Denote  the function $f(x,\ybf) : \Xcal \times \Ycal^Q$ defined by
		$f(x,\ybf) = d(\delta_{\ybf},\mu_x)^2$. Then, with the notations of the Corollary, 
		$D=\Xcal \times \Ycal^Q$ and~:
		
		- $f$ is Borel measurable because of our choice of distance $d$ and the definition of conditional probability distribution;
		
		- both $\Xcal$ and $\Ycal^Q$ are $\sigma$-compact and $f(x,\cdot)$ is continuous thus lower semi-continuous;
		
		- using the Proposition 13 and Remark 14 in~\cite{huber_energy_quantization_turinici22} the set 
		$I=\{ x \in \Xcal : \text{ for some } y \in \Ycal^Q : f(x,y) = \inf f(x,\cdot) \}$ is equal to $\Xcal$;
		
		Then, it follows by the Corollary~1 in \cite[page 904]{brown_measurable_1973} that 
		there exists a measurable function $\ybf^{opt} :\Xcal \to \Ycal^Q$ such that 
		$f(x,\ybf^{opt}(x)) = \inf_{\ybf\in \Ycal^Q} f(x,\ybf)$ i.e. the conclusion.
	\end{proof}

	\begin{remark}
		The procedure described in \cite[pages 906-907]{brown_measurable_1973} 
		allows even to obtain a selection which is compatible with some order relation.
	\end{remark}

	\subsection{Existence of a continuous conditional quantization in 1D}
	We now analyze the continuity of the conditional quantization. We do not have general results but will consider a particular case.
	\begin{proposition}
		Let us take $\Ycal=\R$, $a=0$, $r=1$ (i.e. the kernel is the so-called 'energy' kernel). 
		We work under the assumptions of proposition~\ref{prop:measurable} 
		and suppose in addition that the distribution $\mu(dx,dy)$ is absolutely continuous with respect to the Lebesgue measure and admits a continuous density $\rho(x,y)$ which is strictly positive on $\Xcal \times \Ycal$. Then the conditional quantization $\ybf^{opt}(x)$ is unique for any $x\in \Xcal$ and continuous as a function of $x$.
	\end{proposition}
	\begin{proof}
		Note first that for 
		any fixed but arbitrary $\alpha \in ]0,1[$, the continuity of the density $\rho(x,y)$ implies by standard arguments, the continuity, with respect to $x$, of the quantile $\alpha$ of the law $\mu_x$.
		Using ~\cite[Proposition 21]{huber_energy_quantization_turinici22}, for any $x$ the optimal quantization $\ybf^{opt}(x)$
		is unique and corresponds to the set of quantiles $\frac{q+1/2}{Q}$, 
		$q=0,...,Q-1$
		of the law $\mu_x$. 
		Put together, these facts allow to reach the conclusion.
	\end{proof}
	\begin{remark}
		The hypothesis of absolute continuity of $\mu$ and the hypothesis on the density $\rho(x,y)$ can be weaken (see also~\cite{gannoun2003nonparametric,mehra1991smooth}
		for alternative hypothesis used in this context).   
		
		On the other hand similar results can be proven under more general assumptions; for instance one could  check (proof not given here) that under the assumption that 
		$\sup_{x\neq y, x,y\in \Xcal} \frac{d(\mu_x,\mu_y)}{\|x-y\|^\alpha} < \infty$
		then a Holder-$\alpha$ continuous selection of quantiles can be performed and therefore a Holder-$\alpha$ continuous conditional quantization too.    
	\end{remark}
	
	\begin{remark}
		Another approach could use interpolation: if a quantization is performed for each member of the set of values $x_j$ resulting in vectors $\ybf_j$ then interpolation can be used e.g., for $\Xcal=\R$ and supposing $x_j$ ordered increasingly, for 
		$x= v x_i + (1-v) x_j$ one could combine the vectors $\ybf_j$ with corresponding weights. 
		For higher dimensional $\Xcal$ trilinear interpolation could be invoked.
	\end{remark}

\section{The Deep Conditional Quantization algorithm (DCMQ) : conditional sampling version} \label{sec:algo}

Being now comforted by the theoretical results of the previous section, we look for a practical way to compute the conditional quantization. In particular we will use deep neural networks and will check numerically that such methods can indeed provide good results.

The deep conditional quantization algorithm (abbreviated DCMQ) that we introduce here uses a network that transforms an input $x\in \Xcal$ into a vector $\ybf^{dcmq}(x)\in \Ycal^Q$ with the goal to have  $\ybf^{dcmq}(x)$ as  close as possible to the optimal conditional quantizer $\ybf^{opt}(x)$
of the law $\mu_x$
as  in equation \eqref{eq:defintion_quantizer_by_ineq}.
The procedure is described in algorithm DCMQ below. We describe first the 
default version 
which assumes that a {\bf conditional sampling} is possible, i.e., given $x\in \Xcal$ one can sample from $\mu_x$.
This algorithm will be tested in sections ~\ref{sec:numerics_2Dgaussian_additive}, \ref{sec:numerics_2Dgaussian_std} and \ref{sec:numerics_1Dgaussian_mixture} ; then  
in section~\ref{sec:numerics_mnist_reconstruction} we present the variant 
that samples directly from the joint distribution and use it for the 
restauration of MNIST images. 

\begin{algorithm}
\caption{Deep Conditional Measure Quantization algorithm : DCMQ}
\label{alg:tcalgo}
\begin{algorithmic}[1]
	\Procedure{DCMQ}{}
	\State $\bullet$ set batch size $B$, sampling size $J$, parameters $a$ (default $10^{-6}$), and $r$ (default $1.$), minimization algorithm (default = Adam \cite{kingma_adam_2017}); max iterations (default $1000$)
	\State $\bullet$ choose a network architecture and initialize layers (default~:  $5$ sequential fully connected layers of size $n_y \times Q$, first input is of size $n_x$);
	\While{(max iteration not reached)}
	\State $\bullet$ sample i.i.d $x_1,...,x_B$ according to the marginal law $\mu^X$ of $X$;
	\State $\bullet$ for each $b\le B$  sample i.i.d $J$  times from $\mu_{x_b}$  and denote $\tilde\ybf_b$ the sample as a vector in $\Ycal^J$; 
	
	\State $\bullet$ propagate  $x_1,...,x_B$ through the network to obtain $\ybf^{dcmq}(x_b) \in \Ycal^Q$, $b\le B$
	\State $\bullet$ compute the loss		
	$\Lcal=\frac{1}{B}\sum_{b=1}^B d\left( \delta_{ \tilde\ybf_b}, \delta_{\ybf^{dcmq}(x_b)} \right)^2$;
	\State $\bullet$   update the network as specified by the stochastic
	optimization algorithm	(using backpropagation) to minimize the loss $\Lcal$.	
	\EndWhile
	\EndProcedure
\end{algorithmic}
\end{algorithm}

	The numerical performance of the algorithm is tested below; in all cases when no precision is given the default parameters of the DCMQ algorithm are used.

	\subsection{Quantization of 2D Gaussian conditioned on its mean} \label{sec:numerics_2Dgaussian_additive}

\begin{figure*}[!h]
		\centering
		\includegraphics[width=.24\linewidth]{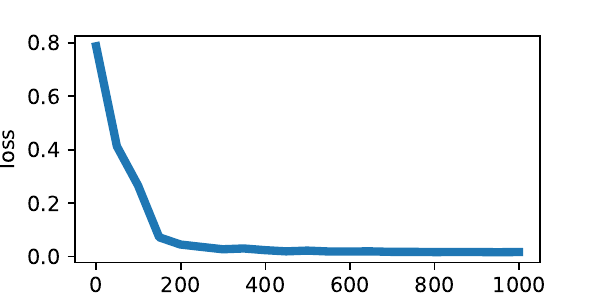}
		\includegraphics[width=.74\linewidth]{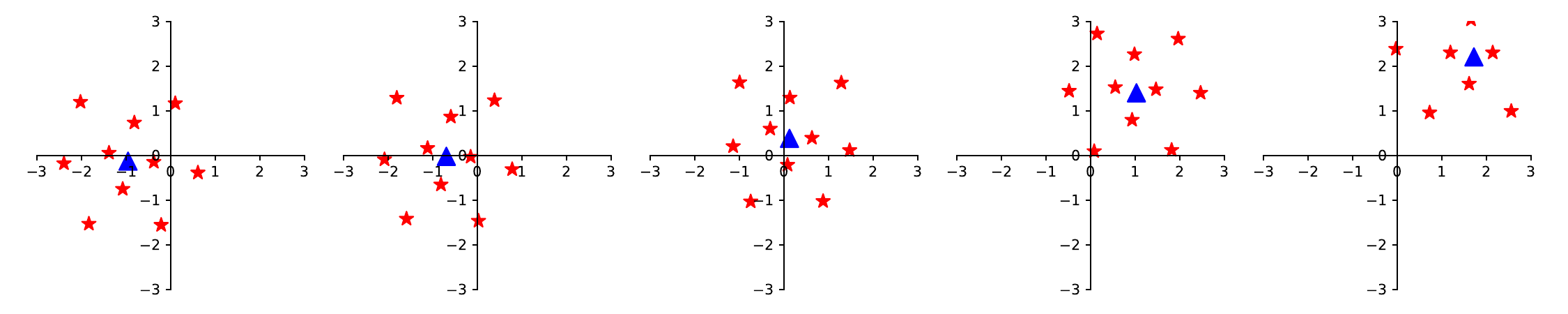}
		\caption{Conditional quantization with $Q=10$ points for the test in section \ref{sec:numerics_2Dgaussian_additive}.
			{\bf Left image:} Convergence of the loss function.
			{\bf Right images~:} Five points  $x_1, ..., x_5\in\Xcal = \R^2$ are sampled from  $\mu^X$ (plotted as blue triangles); the DNN (after training) is asked to quantize	 the conditional distribution 
		  $\mu_{x_b}$ for each $b\le 5$ (red stars). Recall that  $\mu_{x_b}$ is a Gaussian shifted by $x_b$. 
			The quantization points follow precisely the indicated mean.}
		\label{fig:several_quantizations10}
	\end{figure*}

\begin{figure}[!thbp]
		\centering
		\includegraphics[width=.5\linewidth]{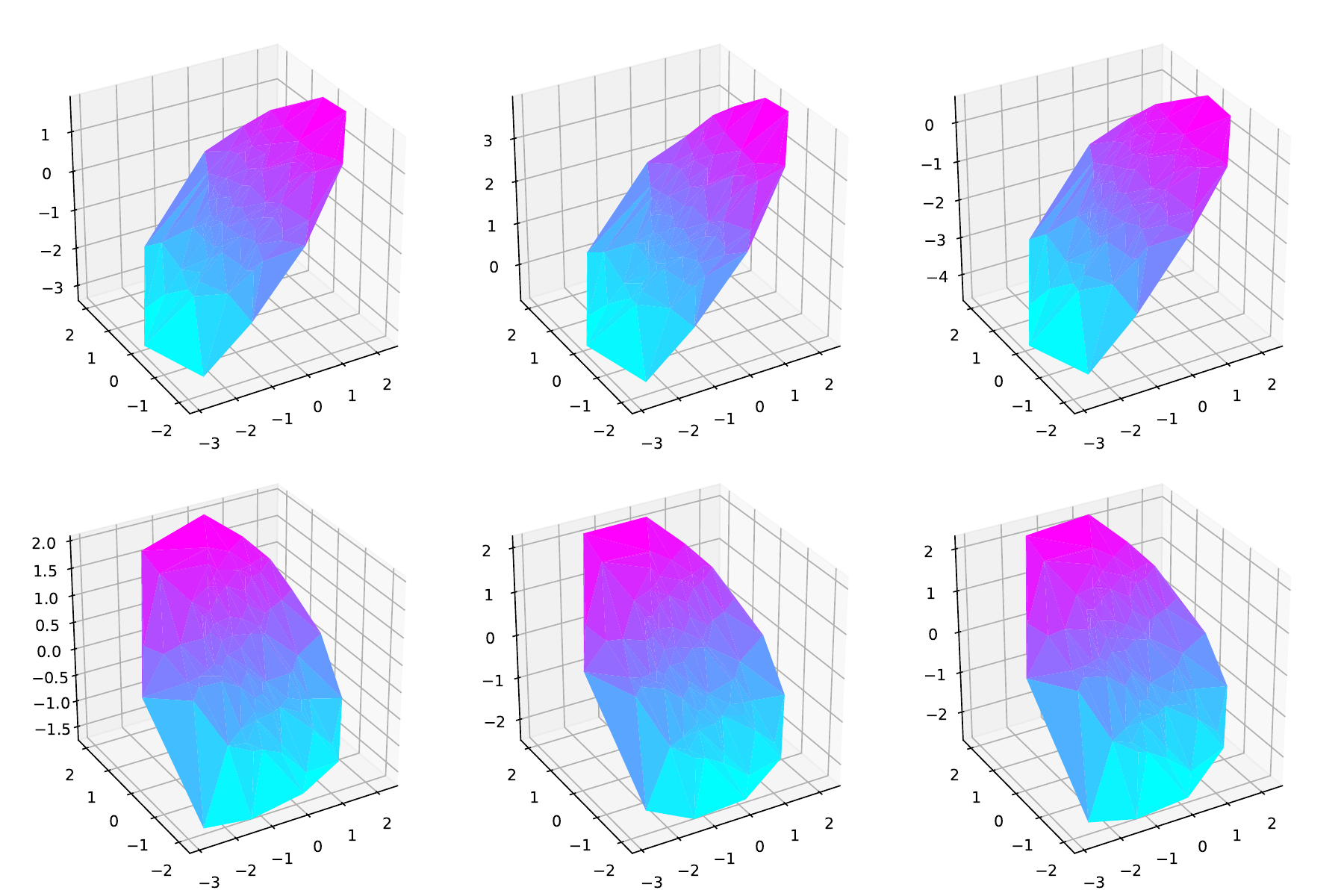}
		\caption{Conditional quantization with $Q=10$ points for the test in section \ref{sec:numerics_2Dgaussian_additive}.
			The conditional quantization points
			$x\in \R^2 \mapsto \ybf^{dcmq}(x)=(\ybf_1^{dcmq}(x),...,\ybf_{10}^{dcmq}(x)) \in (\R^2)^{10}$ are plotted as functions of $x$; each column is a dimension of $\ybf^{dcmq}(x)$ (each row a dimension of the value itself); for graphical convenience  we only plot the first $3$ quantized functions i.e.  
			$\ybf_1^{dcmq}(x)$ (first column), $\ybf_2^{dcmq}(x)$ (second column), $\ybf_3^{dcmq}(x)$ (third column).
			The functions appear smooth and move synchronously, which is a suitable property of the conditional quantization, see section~\ref{sec:theory}.}
		\label{fig:several_quantizations10_conv}
	\end{figure}

	The first test will be a 2D Gaussian that has its mean given by another variable: 
	let $X$ and $Y$ be 2D independent standard Gaussian variables and consider $\mu$ to be the distribution of $X+Y$. The implementation of the DCMQ algorithm is available at  
	\cite{turinici_github_quantization} and the results are presented in figures \ref{fig:several_quantizations10} and \ref{fig:several_quantizations10_conv}.
	The DCMQ algorithm is shown to converge well (cf. figure \ref{fig:several_quantizations10}). Moreover the quantization seems to have desirable properties i.e., it follows the conditional information (the mean). See legend of the figures for additional information.

	\subsection{Quantization of 2D Gaussian : the multiplicative case} \label{sec:numerics_2Dgaussian_std}
	
\begin{figure*}[!htbp]
		\centering
		\includegraphics[width=.24\linewidth]{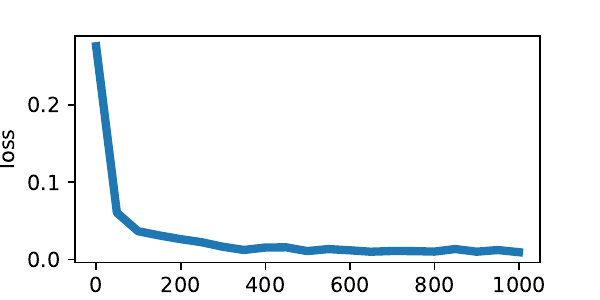}
		\includegraphics[width=.74\linewidth]{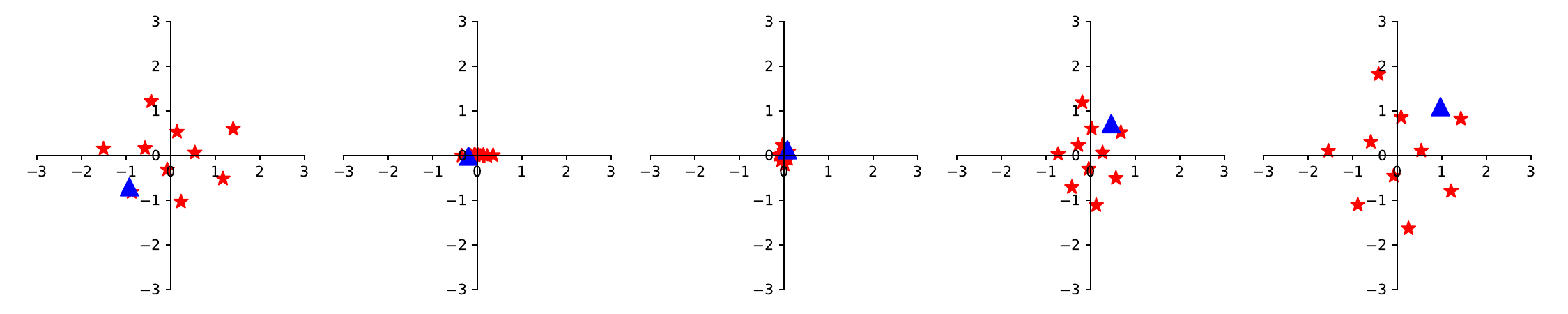}
		\caption{Conditional quantization with $Q=10$ points for the test in section \ref{sec:numerics_2Dgaussian_std}.
			{\bf Left image:} Convergence of the loss function.
			{\bf Right images~:} Five points  $x_1, ..., x_5\in\Xcal = \R^2$ are sampled from  $\mu^X$(blue triangles);  the DNN (after training) is asked to quantize $\mu_{x_b}$ for each $b\le 5$ (red stars). Here 
			$\mu_{x_b}$ is a Gaussian multiplied in each direction by $x_b$. So, for instance when a point $x_b$ has both component values large, the corresponding quantization will look like the quantization of a bi-variate normal. But when $x_b$ is close to some axis, the quantization will act on a very elliptical form distribution because one of the Gaussian is multiplied by a small constant. This expected behavior is reproduced well by the converged DNN.}
		\label{fig:several_quantizations10_std}
	\end{figure*}

\begin{figure}[!htbp]
		\centering
		\includegraphics[width=.5\linewidth]{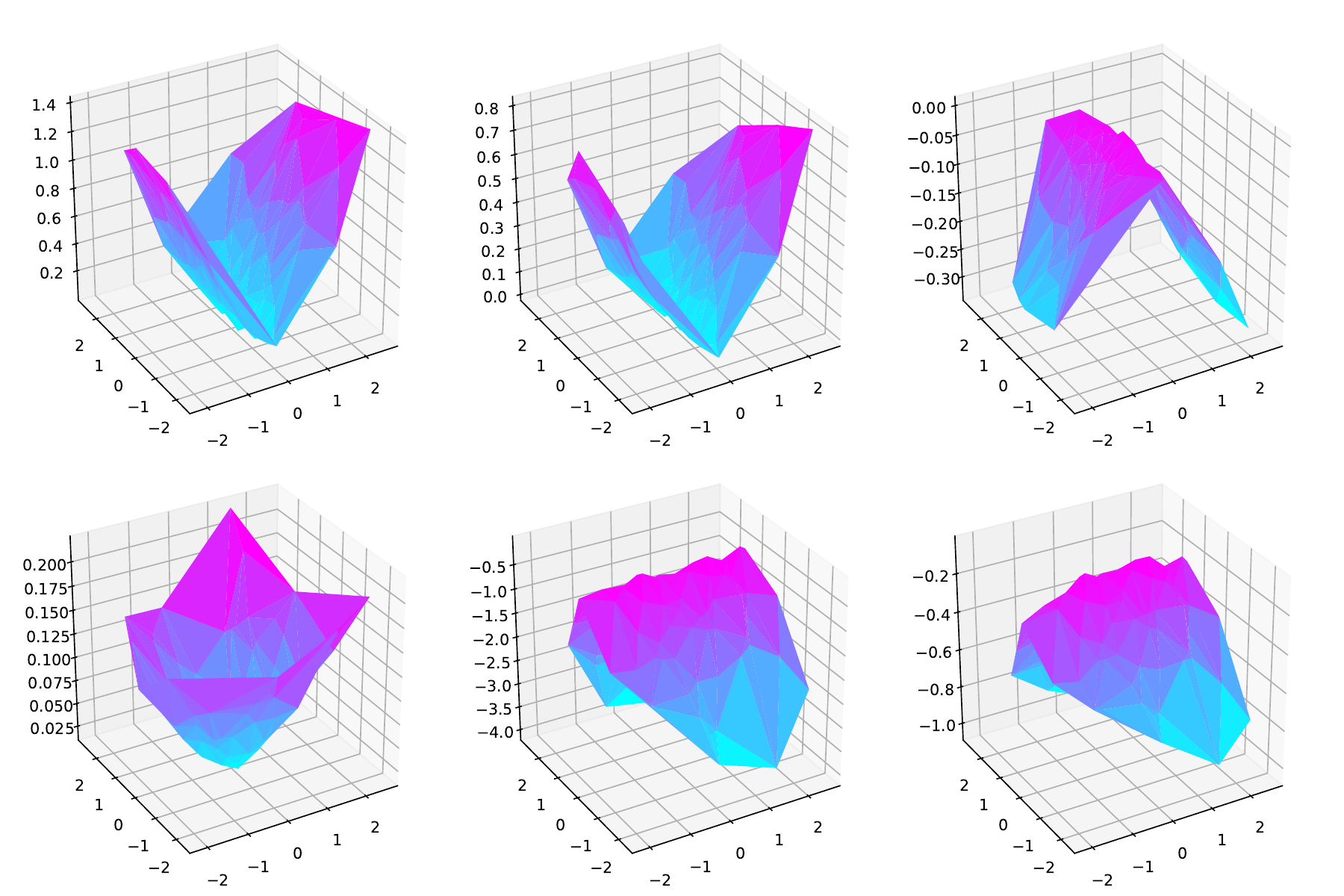}
		\caption{Conditional quantization with $Q=10$ points for the test in section \ref{sec:numerics_2Dgaussian_std}. 
			We plot the conditional quantization  functions $\ybf_1^{dcmq}(x), \ybf_2^{dcmq}(x), \ybf_3^{dcmq}(x)$ as in figure~\ref{fig:several_quantizations10_conv}.}
		\label{fig:several_quantizations10_conv_std}
	\end{figure}

	We move now to another test case where the condition enters multiplicatively; with the notations above ($X$ and $Y$ 2D independent standard Gaussian variables) $\mu$ is taken to be the distribution of $X \cdot Y$. The results are presented in figures  \ref{fig:several_quantizations10_std} and \ref{fig:several_quantizations10_conv_std}.
	The DCMQ algorithms converges well and the quantization follows the conditional information.
	
\subsection{Quantization of a 1D Gaussian mixture crossing} \label{sec:numerics_1Dgaussian_mixture}

\begin{figure}[!htbp]
	\centering
	\includegraphics[width=.33\linewidth]{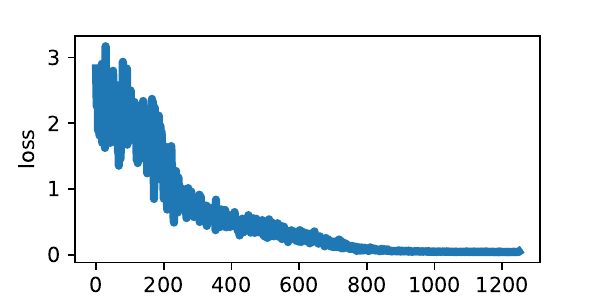}
	
	\includegraphics[width=.95\linewidth]{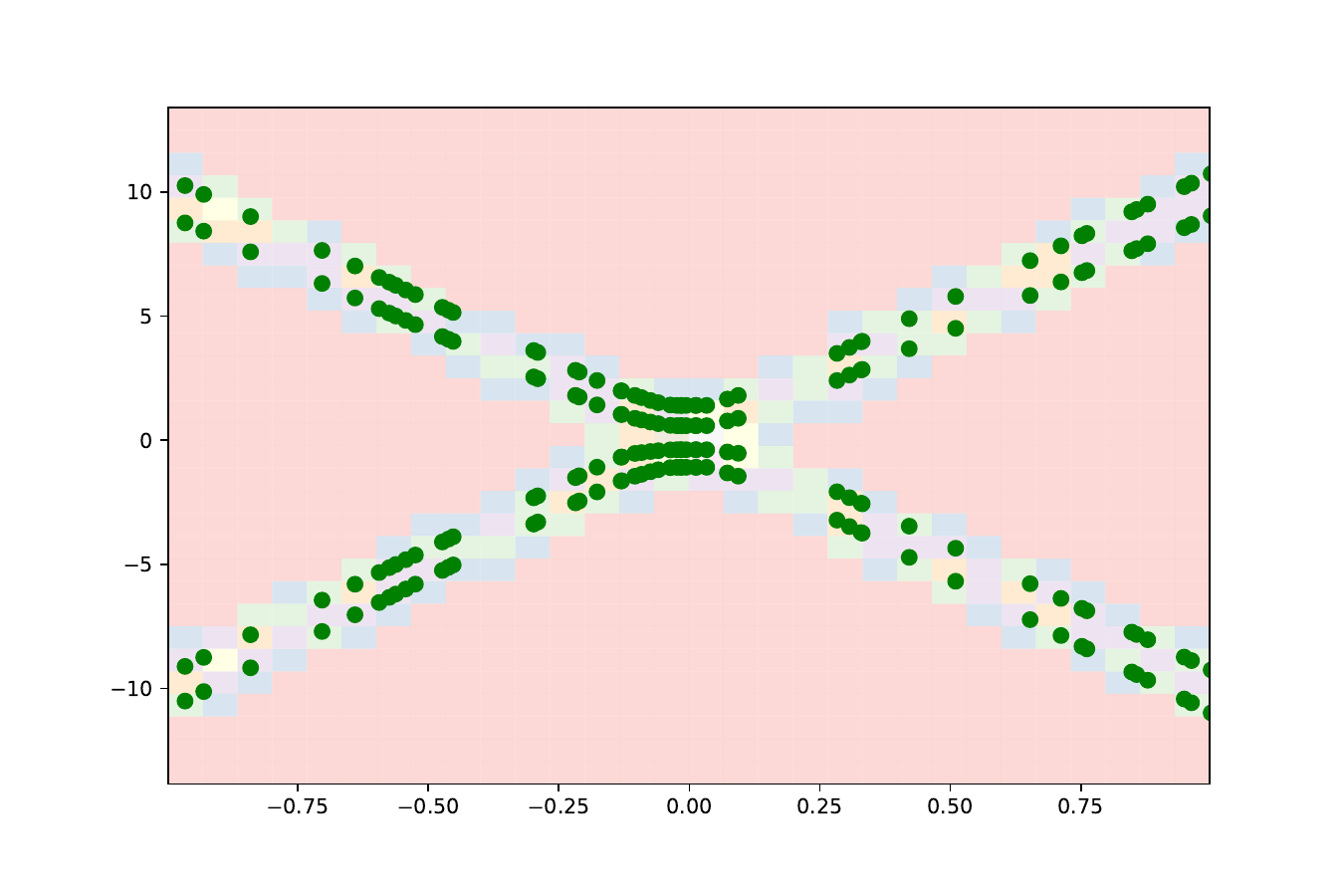}
	\caption{Conditional quantization with $Q=4$ points for the test described in section \ref{sec:numerics_1Dgaussian_mixture}.
		{\bf Top image:} Convergence of the loss function.
		{\bf Bottom image~:} The joint density of $(X,Y)$ is plotted in background (red  are low values, blue are high values). The green dots indicate the quantized values. Note that each of the two parts of the mixture is assigned two quantization points that move along the mean.}
	\label{fig:crossing}
\end{figure}

We consider now the situation of a parameter $X$ uniform in $[-1,1]$ and the dependent variable $Y$ will be a even  mixture of two 1D Gaussians, centered at $\pm 10X$; the joint density of $(X,Y)$ is plotted in figure \ref{fig:crossing} (background).
The DCMQ algorithm converges well  and the quantization follows the expected laws.

\section{The Deep Conditional Quantization through joint sampling : MNIST restauration} \label{sec:numerics_mnist_reconstruction}

Conditional sampling from $\mu_x$ as in section~\ref{sec:algo} is not always possible and data presentation can indicate
joint sampling as the only way to obtain a couple $(X,Y)$~;
this arrives especially when $X$ is continuous and it is impossible to ensure that $X$ has a desired value $x$. 
 We adapt in this section the previous algorithm and test on a image reconstruction task. 

\begin{figure}[!htbp]
	\centering	
	\includegraphics[width=.95\linewidth]{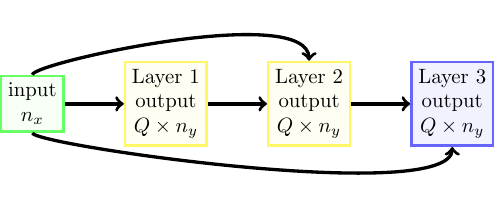}
	\caption{Default network architecture used in section \ref{sec:numerics_mnist_reconstruction} (batch size $B=1$).}
	\label{fig:unet3_architecture}
\end{figure}

The default neural network architecture is as follows~: all layers are fully-connected and have as output a tensor of shape $B \times n_y \times Q$ ($B$ is the batch size)~; the first (input) layer takes data of size $B \times n_x$. All layers act on the output of the previous layer concatenated with the condition (input of the first layer of shape $B \times n_x$). This architecture is akin to a "constant attention" \cite{attention14,vaswani_attention_2017} U-net\cite{shelhamer_fully_2017}.
In our tests the default is to use $3$ such dense layers with ReLU activations. A graphical description is given in figure
\ref{fig:unet3_architecture}.

The algorithm minimizes (by sampling) the following loss functional (with obvious notations)
\begin{equation}
\Ebb_{x \sim \mu^X} \left[ d\left( \mu_x, \frac{1}{Q}\sum_{q=1}^Q \delta_{x,\ybf_q^{dcmq}(\mu_x)} \right)^2 \right].	\label{eq:loss_functional_mux}
\end{equation}

\begin{algorithm}[!htbp]
	\caption{Deep Conditional Measure Quantization algorithm through joint sampling : DCMQ-J}
	\label{alg:tcalgoj}
	\begin{algorithmic}[1]
		\Procedure{DCMQ-J}{}
		\State $\bullet$ set batch size $B$, sampling size $J$, parameters $a$ (default $10^{-6}$), and $r$ (default $1.$), minimization algorithm (default = Adam \cite{kingma_adam_2017}); max iterations (default $1000$)
		\State $\bullet$ choose a network architecture and initialize layers (default : cf. fig. \ref{fig:unet3_architecture});
		\While{(max iteration not reached)}
		\State $\bullet$ sample i.i.d $(x_1,y_1),..., (x_B,y_B)$ from the dataset ;		
		\State $\bullet$ propagate  $x_1,...,x_B$ through the network to obtain $\ybf^{dcmq}(x_b) \in \Ycal^Q$, $b\le B$~;
		\State $\bullet$ compute the loss		
		$\Lcal=\frac{1}{B} \sum_{b=1}^B  d\left( \delta_{x_b,y_b} , 
		\frac{\sum_{q=1}^Q \delta_{x_b,\ybf_q^{dcmq}(x_b)} }{Q}  \right)^2$~;
		\State $\bullet$   update the network as specified by the stochastic
		optimization algorithm (using backpropagation) to minimize the loss $\Lcal$.
		\EndWhile
		\EndProcedure
	\end{algorithmic}
\end{algorithm}

\begin{figure}[!htbp]
	\centering	
	\includegraphics[width=.75\linewidth]{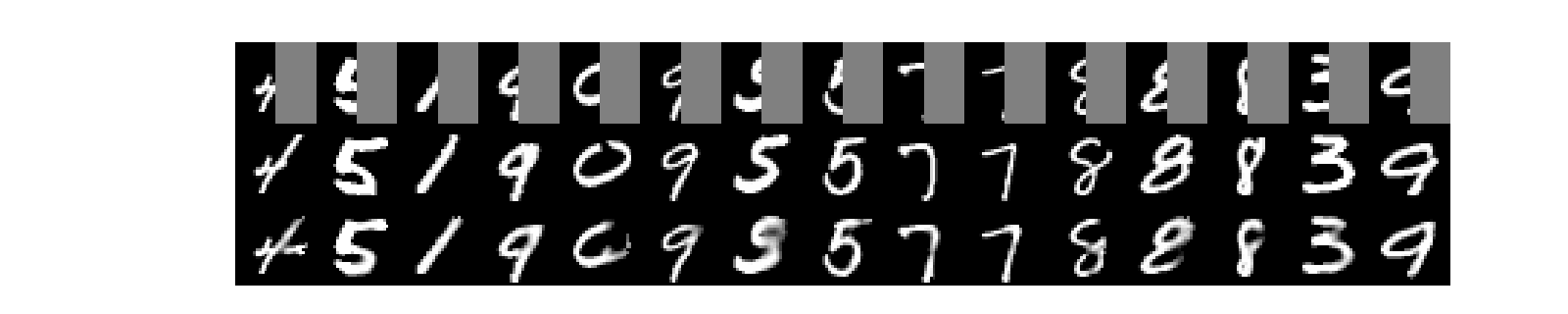}
	\includegraphics[width=.75\linewidth]{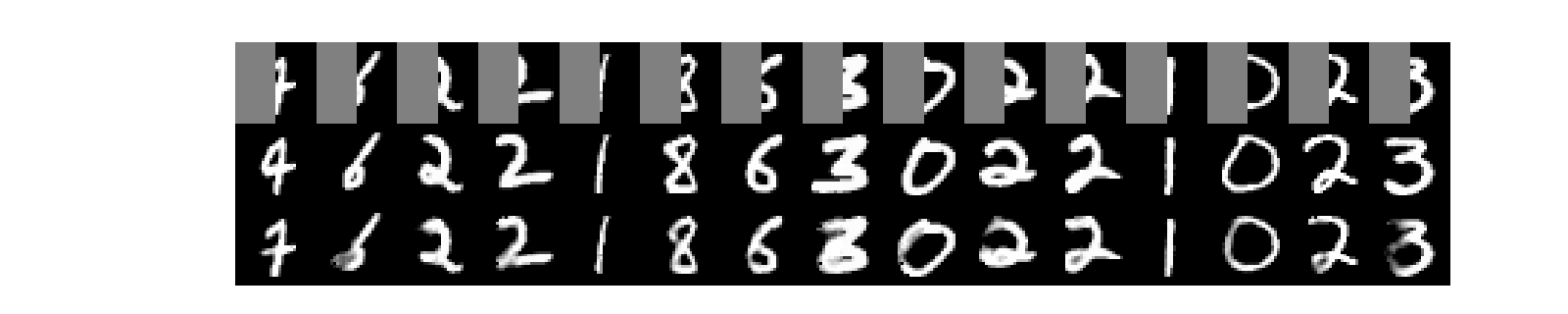}
	\includegraphics[width=.75\linewidth]{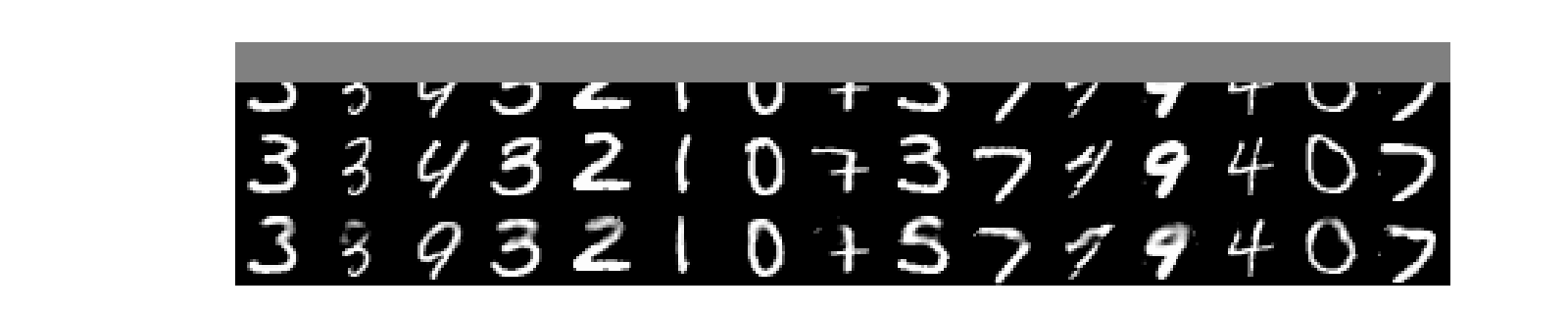}
	\includegraphics[width=.75\linewidth]{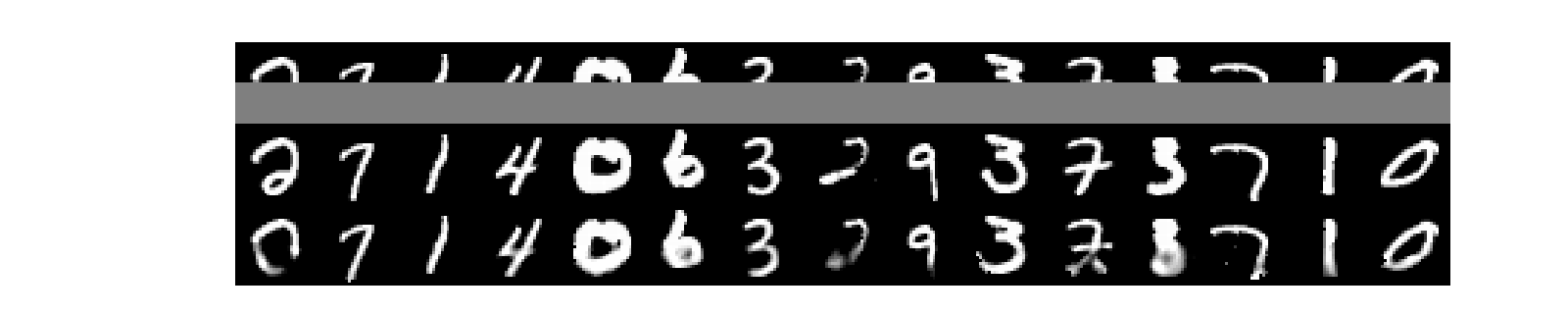}
	\includegraphics[width=.75\linewidth]{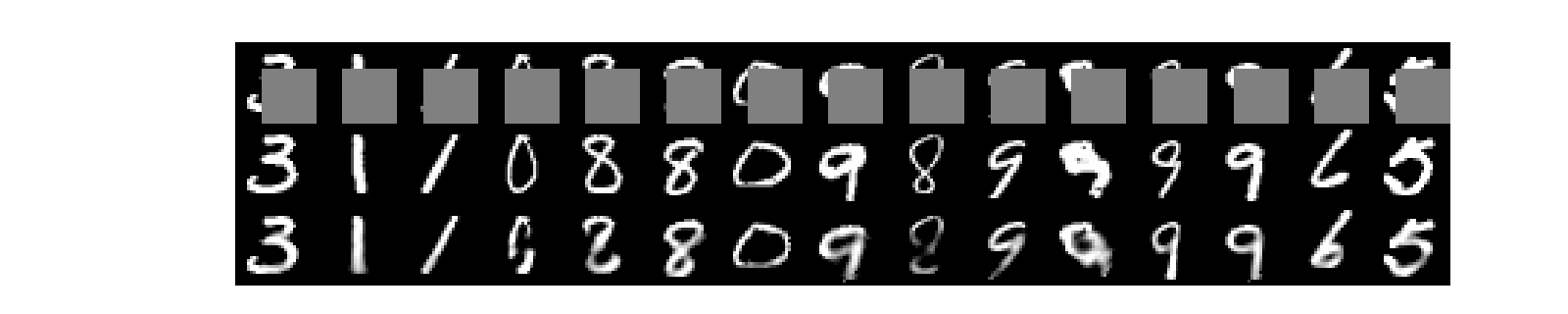}
	\includegraphics[width=.75\linewidth]{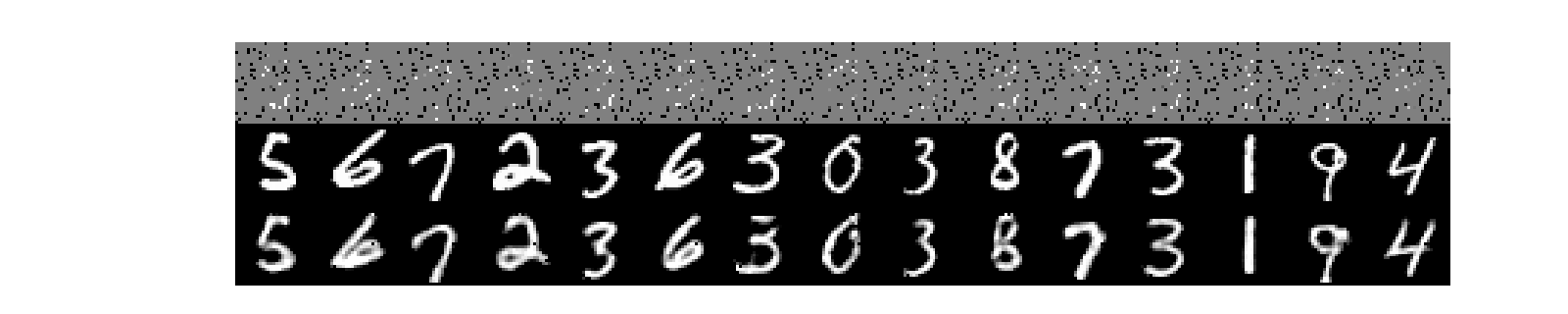}
	\caption{MNIST reconstruction results as described in section~\ref{sec:numerics_mnist_reconstruction}.}
	\label{fig:mnist_reconstruction}
\end{figure}

\begin{remark}
Note that an alternative formulation can be proposed that employs the following loss functional instead of \eqref{eq:loss_functional_mux}~:
\begin{equation}
d\left( \mu, \frac{1}{Q}\sum_{q=1}^Q \delta_{x,\ybf_q^{dcmq}(\mu_x)} \right)^2.	\label{eq:loss_functional_mu_joint}
\end{equation}
This loss has lower variance (gain of order $O(\frac{1}{B})$) but requires more computations per iteration~: $O(B^2 (1+Q)^2 n_y +  B^2 n_x)$ instead of $O(B Q^2 n_y)$. 
The sampled version used by the algorithm is in this case~:
		$ d\left( \frac{\sum_{b=1}^B \delta_{x_b,y_b}}{B} , 
\frac{\sum_{b=1}^B \sum_{q=1}^Q \delta_{x_b,\ybf_q^{dcmq}(x_b)} }{B \cdot Q }  \right)^2$.
\end{remark}

The algorithm was tested on the MNIST  dataset for an inpainting (image restoration) task 
i.e., it was requested to restore the whole image using only a part of it as input~; 
in all cases we considered a part of the image as known corresponding to the condition i.e., the $X$ component (of dimension $n_x$), and another part as unknown, corresponding to the $Y$ component (of dimension $n_y$)~; the $Q=1$ quantization was learned using the {\bf train} dataset and, at test time, we asked the algorithm to output the quantization for images in the {\bf test} MNIST dataset. Several situations were considered for the unknown part $Y$  depicted in gray
in figure~\ref{fig:mnist_reconstruction} (the description is consistent with the order of results there)~:
right half, left half,
  upper half,  lower half,
lower right $2/3 \times 2/3$ corner, 
random pixels covering $90\%$ of all $28\times 28$ pixels. The known part,  plotted in the first row, was given as input ; the true images are in the second row and the guess of the algorithm is in the third row. Good agreement is observed in all test situations as the algorithm manages to reconstruct well the missing parts.
	
\section{Concluding remarks} \label{sec:concluding}
	
In this paper we discuss the quantization of conditional probability laws. We first prove that in general the quantization can be represented as measurable function and then, in a particular case, we give a theoretical result that ensures the quantization points depend continuously on the condition. Then, a deep learning algorithm is introduced using Huber-energy kernels to find numerically the solution of the quantization problem. The procedure is tested on some standard cases and shows promising results.

\bibliographystyle{kr}
\bibliography{deep_conditional_measure_quantization}
	
\end{document}